\documentclass[conference]{IEEEtran}
\IEEEoverridecommandlockouts
\usepackage{cite}
\usepackage{amsmath,amssymb,amsfonts}
\usepackage[a4paper, margin=1.8cm]{geometry}
\usepackage{algorithmic}
\usepackage{fancyhdr}
\usepackage{graphicx}
\usepackage{lipsum}
\usepackage{textcomp}
\usepackage{verbatim}
\usepackage{multirow}
\usepackage{array}
\usepackage{float}
\usepackage{xcolor}
\usepackage{hyperref}
\usepackage{enumitem}
\usepackage{subcaption} 
\usepackage{balance}

\def\BibTeX{{\rm B\kern-.05em{\sc i\kern-.025em b}\kern-.08em
    T\kern-.1667em\lower.7ex\hbox{E}\kern-.125emX}}
\usepackage{hyperref}
\usepackage{geometry}

\fancypagestyle{firstpage}
{
    \fancyhead[R]{} 
    \fancyhead[L]{2024 27th International Conference on Computer and Information Technology (ICCIT),\\ 20-22 December 2024, Cox’s Bazar, Bangladesh}
    \fancyfoot[C]{} 
    \fancyfoot[L]{979-8-3315-1909-4/24/\$31.00 ©2024 IEEE}
    \setlength{\footskip}{25pt} 
}

\begin{document}

\title{Empowering Bengali Education with AI: Solving Bengali Math Word Problems through Transformer Models}

\author{
    \IEEEauthorblockN{
        Jalisha Jashim Era \IEEEauthorrefmark{1}, 
        Bidyarthi Paul \IEEEauthorrefmark{2}, 
        Tahmid Sattar Aothoi \IEEEauthorrefmark{3},
        Mirazur Rahman Zim \IEEEauthorrefmark{4},\\
        Faisal Muhammad Shah \IEEEauthorrefmark{5}
    }

    \IEEEauthorblockA{
        \IEEEauthorrefmark{1}  \IEEEauthorrefmark{2}  \IEEEauthorrefmark{3}  \IEEEauthorrefmark{4} \IEEEauthorrefmark{5} Department of Computer Science and Engineering
    }
    \IEEEauthorblockA{
        \IEEEauthorrefmark{1}  \IEEEauthorrefmark{2}  \IEEEauthorrefmark{3} \IEEEauthorrefmark{4} \IEEEauthorrefmark{5} Ahsanullah University of Science and Technology, Dhaka, Bangladesh \\
    }
    \IEEEauthorblockA{
        Emails: \{\IEEEauthorrefmark{1}ira16jalisa, \IEEEauthorrefmark{2}bidyarthipaul01,
        \IEEEauthorrefmark{3}tahmidaothoi007,
        \IEEEauthorrefmark{4}miraz.zim.38\}@gmail.com, \IEEEauthorrefmark{5}faisal.cse@aust.edu
    }
}

\maketitle

\thispagestyle{firstpage}

\begin{abstract}

Mathematical word problems (MWPs) involve the task of converting textual descriptions into mathematical equations. This poses a significant challenge in natural language processing, particularly for low-resource languages such as Bengali. This paper addresses this challenge by developing an innovative approach to solving Bengali MWPs using transformer-based models, including Basic Transformer, mT5, BanglaT5, and mBART50. To support this effort, the ``PatiGonit" dataset was introduced, containing 10,000 Bengali math problems, and these models were fine-tuned to translate the word problems into equations accurately. The evaluation revealed that the mT5 model achieved the highest accuracy of 97.30\%, demonstrating the effectiveness of transformer models in this domain. This research marks a significant step forward in Bengali natural language processing, offering valuable methodologies and resources for educational AI tools. By improving math education, it also supports the development of advanced problem-solving skills for Bengali-speaking students.\\

\textit   {Keywords-} Math Word Problems, Low-resource Language, Transformer, multilingual
\end{abstract}

\section{\textbf{Introduction}}

The challenge of solving natural language math word problems (MWPs) has intrigued researchers since the inception of artificial intelligence. These problems, often encountered by elementary students, require the conversion of textual descriptions into mathematical equations involving basic arithmetic operations such as addition, subtraction, multiplication, and division. Automating this process using advanced natural language processing (NLP) techniques can significantly enhance educational tools. 

Since the 1960s, researchers have sought to teach computers to solve these kinds of math problems just like humans~\cite{feigenbaum2003some}. With recent technological advancements, we have made significant progress, particularly for simpler problems. Despite advancements in NLP and machine learning, there has been a notable gap in applying these technologies to the Bengali language. Bengali, being a low-resource language, lacks substantial computational resources and datasets to support extensive research and development in NLP applications. Addressing this gap, our study introduces an innovative approach to solving Bengali MWPs using transformer-based models, including the Basic Transformer~\cite{vaswani2017attention}, mT5~\cite{mt5m}, BanglaT5~\cite{bt5}, and mBART50~\cite{mbart50}. These models have been trained on a specifically curated dataset, ``PatiGonit" means ``Arithmetic" or ``Elementary Mathematics".

\begin{figure}[h]
\centering
\includegraphics[width=0.30\textwidth]{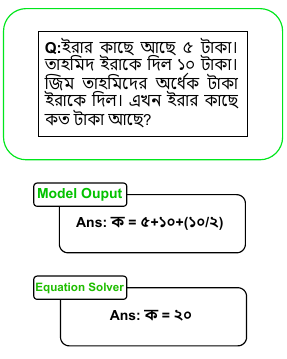} 
\caption{Bengali Math Word Problem Equation generation and solver example}
\label{fig:example}
\end{figure}

For our study, we employed transformer-based models to identify mathematical equations embedded within the text of Bengali word problems~\cite{nayak80mathbot}. We fine-tuned key hyperparameters, including learning rate, number of epochs, and batch size, to optimize model performance. Our approach involves predicting the equation from the word problem using deep learning-based natural language processing techniques. Once the equation is predicted, we use an equation solver to obtain the final answer. An example of the type of math word problem handled by our work is shown in Figure~\ref{fig:example}. The intermediate equations generated by the models are processed by our equation solver to produce the final solution.

In summary, our research advances the state-of-the-art in MWP solving for low-resource languages, providing a robust framework for future studies. This work aims to make educational technology more accessible by filling the gap in multilingual and interpretable math word problem solvers. It also supports the development of critical thinking skills and encourages data-driven decision making in education. Our key contributions include:

\begin{itemize}
    \item \textbf{Creation of the ``PatiGonit" Dataset:} We introduce the ``PatiGonit" dataset, consisting of 10,000 Bengali Math Word Problems from elementary school. This dataset addresses a significant gap in Bengali computational research, providing a valuable resource for developing educational tools tailored to Bengali-speaking students.

    \item \textbf{Application and Fine-tuning of Transformer-based Models:} We implement and fine-tune several transformer-based models, including Basic Transformer, mT5, BanglaT5, and mBART50, specifically for solving Bengali Math Word Problems. Through a comparative analysis, we determine the effectiveness of these models for Bengali Math Word Problem solving.

    \item \textbf{Advancement in Bengali Natural Language Processing (NLP):} We advance Bengali NLP by using transformer-based models to accurately translate and process Bengali math word problems into mathematical equations. This approach improves model accuracy and broadens the application of NLP in educational contexts for Bengali learners.
\end{itemize}

\section{\textbf{Related Works}}
Multilingual models enable the processing and generation of text across multiple languages, enhancing various NLP tasks. Niyarepola et al.~\cite{mt5} employ these models to generate math word problems (MWPs) through a sequence-to-sequence framework with cross-lingual evaluation, adapting datasets like Math23K and MathQA for bilingual assessment. Luo et al.~\cite{llm} introduce MATHWELL, a multilingual model fine-tuned to generate math problems for K-8 students, focusing on generating solvable and accurate questions. Both papers highlight how multilingual models can enrich MWP generation, improving mathematical education across languages.

Several studies like Zhang et al.~\cite{7} and Wang et al.~\cite{8} have explored Seq2Seq models for neural machine translation and solving math word problems (MWPs). In the paper of Zhang et al.~\cite{7},the VNMT model employs LSTM networks for both the encoder and decoder without any pretrained models, achieving a BLEU score of 32.07 for Chinese-English translation and 19.58 for English-German translation, and an accuracy of 79.8\% on the MAWPS single operation dataset. Wang et al.~\cite{8} introduced a Seq2Seq model using a hybrid approach with a GRU encoder and LSTM decoder, achieving 64.7\% on the Math23K dataset and 59.5\% on the MAWPS dataset. The Saligned model of Chiang and Chen~\cite{12}, a Seq2Seq neural encoder-decoder architecture, incorporates LSTM and Graph Convolutional Networks as the encoder and a TreeDecoder, achieving over 65\% accuracy on the Math23K dataset. Graph2Tree, which also uses an LSTM with Graph Convolutional Networks and a TreeDecoder, addresses poor quantity representations and achieved 77.4\% on MAWPS and 69.5\% on Math23K .

Recent research of Liang et al.~\cite{15} and Raiyan et al.~\cite{16} focuses on improving contextual and semantic understanding using transformer-based models. In the study of Liang et al.~\cite{15} MWPBert, which utilizes a pretrained BERT as the encoder and a TreeDecoder, achieved 84.7\% on the Math23K dataset. DeBERTa of Raiyan et al.~\cite{16}, combined with an Enhanced Mask Decoder and a Voting Mechanism, reached 91.0\% on MAWPS and 79.1\% on PARAMAWPS.

Hybrid models that combine different approaches have shown significant improvements. An ensemble model by Wang et al.~\cite{9} utilizing BiLSTM and LSTM with equation normalization achieved 69.2\% accuracy on the Math23K dataset . In the paper Xie and Sun~\cite{10} the GTS model, integrating GRU, TreeDecoder, and gated feedforward networks, achieved 74.3\% on Math23K and 83.5\% on the MAWPS single operation dataset. WARM has been introduced in Chatterjee et al.~\cite{17} and it uses a weakly supervised approach with a bidirectional GRU encoder and three fully connected networks as the decoder, achieving 66.9\% on All Arith and 56.0\% on Math23K .

Unique methodologies have been proposed in Qin et al.~\cite{14} to tackle specific challenges in MWPs. SAUSolver in paper of Qin et al.~\cite{14} introduced a Universal Expression Tree (UET) and auxiliary tasks, achieving up to 62.03\% on the HMWP dataset.

\section{\textbf{Dataset}}

\subsection{Dataset Collection}
Prior to this study, there were no available datasets comprising Bengali math word problems, a gap that significantly limited computational research and the development of educational tools for Bengali-speaking students. To address this void, we introduced the Bengali Math Word Problems dataset named ``PatiGonit", comprising 10,000 math word problems. It traditionally refers to the branch of mathematics dealing with basic numerical operations such as addition, subtraction, multiplication, and division. These problems, initially in English within the MAWPS~\cite{patel2021nlp} dataset, which encompasses a total of 38,000 problems, underwent the translation process for the conversion into Bengali. Due to resource constraint we could only analyzed \textbf{10,000} of these problems. The dataset, typical of elementary school-level math, includes both basic arithmetic and some algebraic challenges, organized into two columns: one for the word problems and the other for the corresponding mathematical equations.

We present an analysis of the equations contained within the dataset, categorizing them into simple and complex equations based on the number of mathematical symbols used. Simple equations are defined as those containing only one mathematical operation, while complex equations involve multiple operations.The Table~\ref{tab:count} below summarizes the total count of simple and complex equations, as well as a further breakdown of the simple equations into four categories: addition, subtraction, multiplication, and division. 

\begin{table}[H]
\caption{Summary of Equation Types in our proposed dataset}
\label{tab:count}
\tiny
\resizebox{\columnwidth}{!}{%
\begin{tabular}{ll r}
\hline
\multicolumn{2}{c}{\textbf{Equation Types}}                            & \textbf{Count} \\ \cline{1-3}
\multicolumn{1}{c}{\multirow{4}{*}{Simple Equations}} & Addition (+)       & 1761           \\  
\multicolumn{1}{c}{}                                  & Subtraction (-)    & 3217           \\ 
\multicolumn{1}{c}{}                                  & Multiplication (*) & 1610           \\ 
\multicolumn{1}{c}{}                                  & Division (/)       & 3164           \\ \hline
\multicolumn{1}{c}{Complex Equations}                 & Mixed (+,-,*,/)          & 248            \\ \hline
\end{tabular}%
}
\end{table}

\subsection{Dataset Annotation}

\begin{figure}[h]
\centering
\includegraphics[width=1.0\columnwidth]{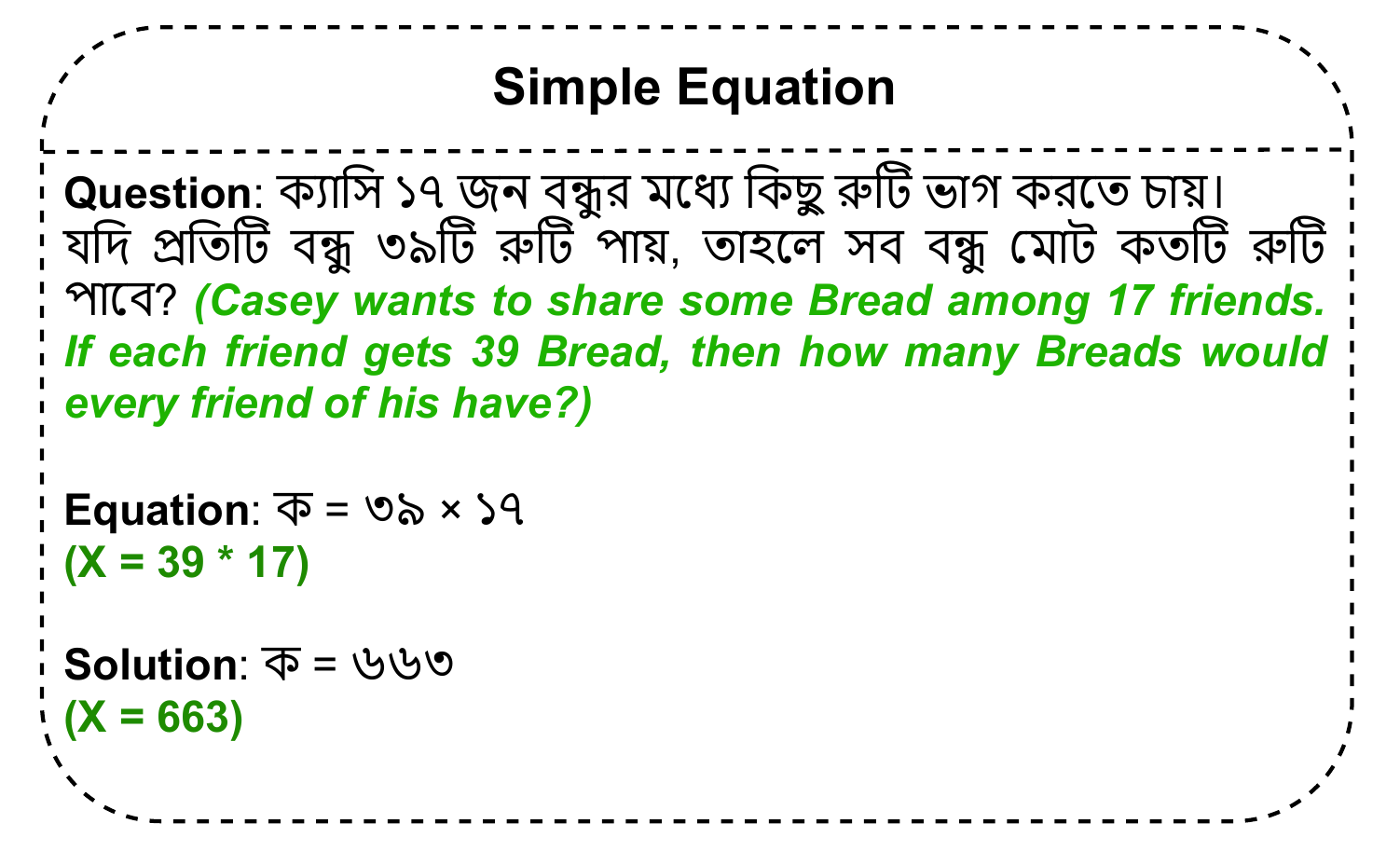} 
\caption{Sample of a simple equation}
\label{fig:sam1}
\end{figure}

\begin{figure}[h]
\centering
\includegraphics[width=1.0\columnwidth]{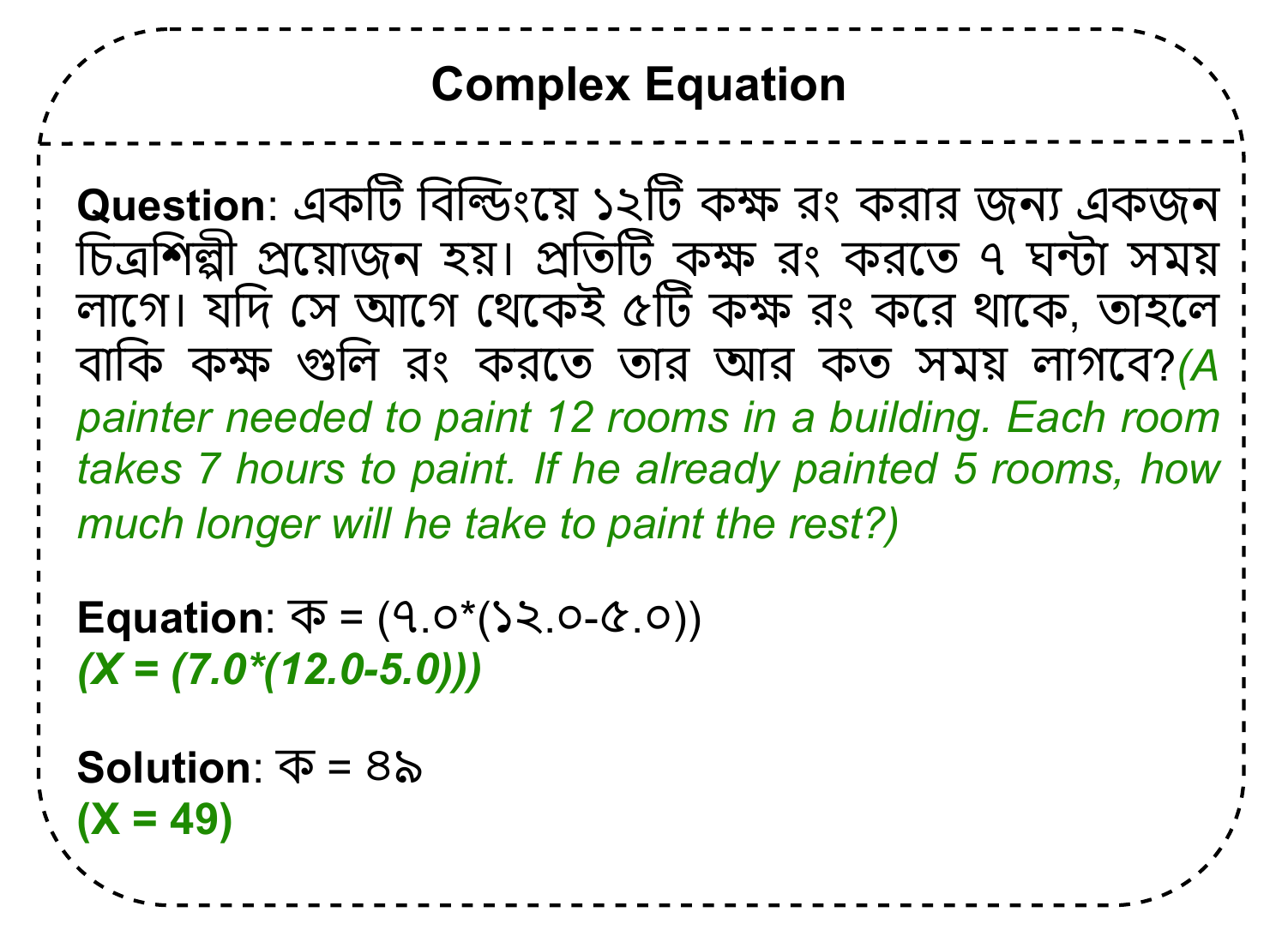} 
\caption{Sample of a complex equation}
\label{fig:sam2}
\end{figure}

For our dataset of math word problems, we, the native speakers meticulously reviewed and annotated each problem to ensure they were accurate and suitable for local use. This involved:
\begin{itemize}
    \item Translating the math word problems into Bengali.
    \item Adjusting numbers to fit Bengali formats.
    \item Changing currency symbols to match local standards.
    \item Revising specific terms, especially in mathematical equations, to ensure they were correct in Bengali.
\end{itemize}
We made these adjustments ourselves to ensure the dataset was not only translated accurately but also culturally relevant and practical for use in education within Bengali-speaking regions. Figure~\ref{fig:sam1} and Figure~\ref{fig:sam2}, shows us a sample for both the simple and complex equation present in our proposed dataset.

Figure~\ref{fig:prob} represents our ``PatiGonit" dataset, alongside their English translations, highlighting key challenges in solving them. Each problem presents a mathematical scenario, and the "Challenges" column points out specific linguistic or cognitive difficulties, such as pronoun usage, sentence structure, or identifying key words. The table effectively showcases how variations in language expression impact problem-solving, providing insights into both language processing and mathematical understanding.

\subsection{Dataset Pre-processing}
Pre-processing refines raw data for Transformer Based Models. This phase begins by lowercasing all input strings and removing leading and trailing spaces. This is necessary to standardize text and reduce whitespace. We then use regular expressions to correctly space punctuation marks, including the Bengali danda. Punctuation is treated as tokens, which helps the model recognize and process textual elements.

\begin{figure*}[H]
\centering
\includegraphics[width=0.8\textwidth]{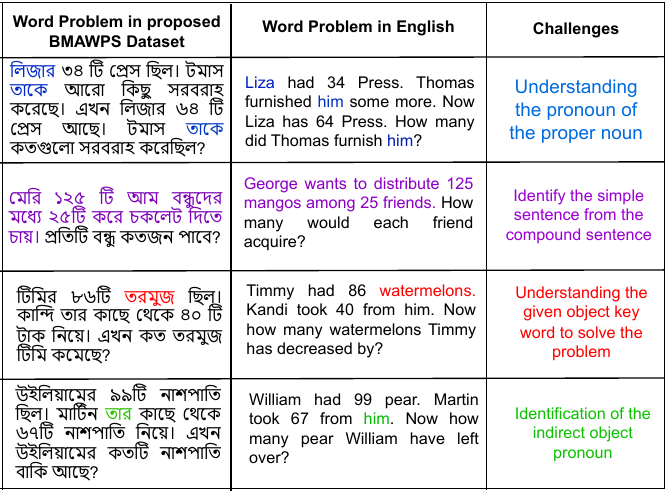} 
\caption{Some challenges of the Bengali Word Problems}
\label{fig:prob}
\end{figure*}

\subsection{Train-Test-Validation Splits}
We split the dataset into three parts: 80\% for training, 10\% for validation, and 10\% for testing. This 80:10:10 split offers a well-balanced approach, providing ample data for the model to learn effectively while retaining a sufficient validation set for tuning hyperparameters and a robust test set for evaluating performance on unseen data.

\begin{figure}[h]
\centering
\includegraphics[width=1.0\columnwidth]{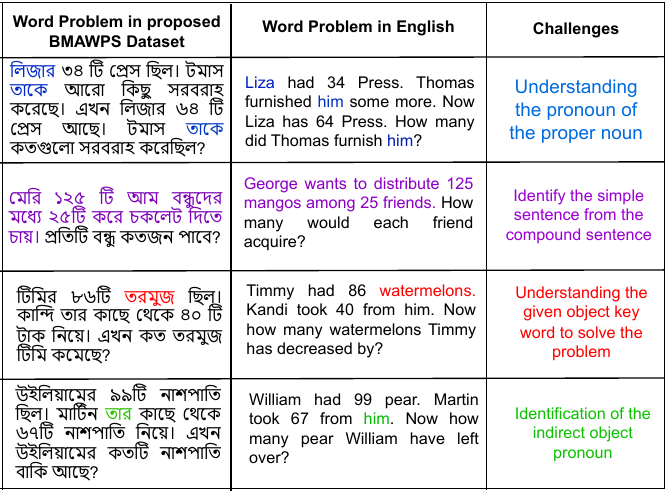} 
\caption{Some challenges in solving the Bengali Math Word Problems}
\label{fig:prob}
\end{figure}

\section{\textbf{Methodology}}

Our methodology centers on using transformer-based models—MT5, BanglaT5, and mBART50—to translate and process Bengali math word problems into mathematical equations. These models were selected over traditional deep learning architectures like BERT, LSTM, and RNN due to their superior ability to handle complex sequence-to-sequence tasks.

\begin{table*}[h]
\renewcommand{\arraystretch}{1.5}
\centering
\caption{Architectural comparison of Transformer and Traditional Deep Learning Models}
\label{tab:archi_comp}
\scalebox{0.92}{%
\begin{tabular}{l c c c c c c}
\hline
\textbf{Feature}         & \textbf{BERT}                                                            & \textbf{LSTM}                                                                  & \textbf{RNN}                                                                   & \textbf{mT5}                                                          & \textbf{BanglaT5}                                                     & \textbf{mBART50}                                                             \\ \hline
\textbf{Architecture}    & \begin{tabular}[c]{@{}c@{}}Encoder-only \\ (bi-directional)\end{tabular} & \begin{tabular}[c]{@{}c@{}}Recurrent, \\ with memory cells\end{tabular}        & \begin{tabular}[c]{@{}c@{}}Recurrent, \\ simpler than LSTM\end{tabular}        & \begin{tabular}[c]{@{}c@{}}Encoder-Decoder \\ (Seq2Seq)\end{tabular}  & \begin{tabular}[c]{@{}c@{}}Encoder-Decoder \\ (Seq2Seq)\end{tabular}  & \begin{tabular}[c]{@{}c@{}}Encoder-Decoder \\ (Seq2Seq)\end{tabular}         \\ \hline
\textbf{Task Type}       & \begin{tabular}[c]{@{}c@{}}Text classification, \\ NLU\end{tabular}      & \begin{tabular}[c]{@{}c@{}}Sequential tasks \\ (time series, NLP)\end{tabular} & \begin{tabular}[c]{@{}c@{}}Sequential tasks \\ (time series, NLP)\end{tabular} & \begin{tabular}[c]{@{}c@{}}Text generation, \\ NLG + NLU\end{tabular} & \begin{tabular}[c]{@{}c@{}}Text generation, \\ NLG + NLU\end{tabular} & \begin{tabular}[c]{@{}c@{}}Text generation, \\ NLG + NLU\end{tabular}        \\ \hline
\textbf{Training Corpus} & \begin{tabular}[c]{@{}c@{}}Wikipedia, \\ BookCorpus, etc.\end{tabular}   & Task-specific                                                                  & Task-specific                                                                  & \begin{tabular}[c]{@{}c@{}}mC4 \\ (multilingual)\end{tabular}         & \begin{tabular}[c]{@{}c@{}}Bengali \\ language corpus\end{tabular}    & \begin{tabular}[c]{@{}c@{}}CC25, \\ Multilingual (50 languages)\end{tabular} \\ \hline
\end{tabular}%
}
\end{table*}

\begin{figure*}[h]
\centering
\includegraphics[width=1.0\textwidth]{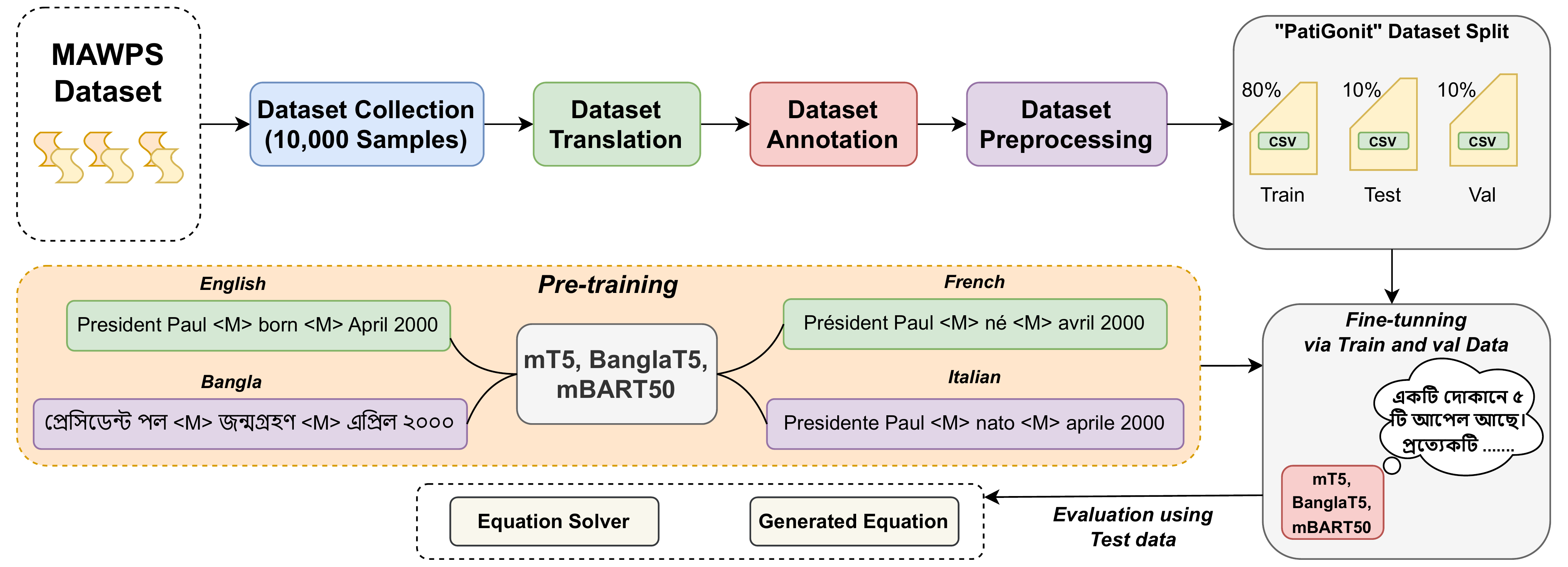} 
\caption{Schematic diagram of Bengali Math Word Problem solver}
\label{fig:meth}
\end{figure*}

Transformers effectively manage long-range dependencies in text, crucial for solving math word problems. Unlike LSTMs and RNNs, which struggle with longer sequences, transformers process entire sequences in parallel, capturing context more efficiently for accurate equation generation. While BERT excels at understanding text, its encoder-only architecture makes it less suited for generative tasks like equation generation. By fine-tuning pretrained versions of mT5, BanglaT5, and mBART50 on our ``PatiGonit" dataset, we optimized the models to excel at generating accurate mathematical equations from Bengali text. mT5 was trained on 101 languages, mBART50 on 50 languages, and BanglaT5 was specifically trained on Bengali. In Figure~\ref{fig:meth}, this is illustrated by providing examples in English, Bengali, French, and Italian, demonstrating the models' multilingual capabilities in understanding and processing text across different languages. Figure~\ref{fig:meth} presents the schematic diagram of the Bengali Math word problem solver.

 To better understand the differences between the selected transformer models (MT5, BanglaT5, mBART50) and other deep learning architectures (BERT, LSTM, RNN), Table \ref{tab:archi_comp} provides a comparative analysis. This table highlights the variations in architecture, task type suitability, and training corpus, which illustrates why transformer-based models are better suited for our task of translating math word problems into mathematical equations. An overview of the models are given below.

\subsection{Basic Transformer}

The complexity of math word problems poses a significant challenge in natural language processing, particularly in converting text to mathematical equations. To tackle this, we employed the Basic Transformer model, renowned for its success in sequence-to-sequence tasks. Developed by~\cite{vaswani2017attention}, the transformer architecture processes entire texts simultaneously, enhancing context comprehension, which is crucial for accurately generating corresponding equations from word problems.

In the transformer-based model~\cite{nayak80mathbot}, the encoder captures semantic relationships in the input text, converting it into a high-dimensional space, while the decoder generates the output equation, guided by the encoder's output and previous predictions. The attention mechanism ensures accuracy by dynamically focusing on relevant parts of the input text, enhancing the model's performance on our Bengali Math Word Problem dataset. Below are the important functions of transformers taken from~\cite{nayak80mathbot}.

\textbf{Attention Function}

\begin{equation}
\text{Attention}(Q, K, V)=\text{Softmax}\left(\frac{QK^T}{\sqrt{d_k}}\right)V
\end{equation}
\begin{equation}
\text{MultiHead}(Q, K, V)=\text{Concat}(\text{head}_1,..., \text{head}_h)W^O
\end{equation}
\begin{center}
where, $\text{head}_i=\text{Attention}(QW_i^Q, KW_i^K, VW_i^V)$
\end{center}

\subsection{mT5}
The mT5 is a multilingual variant of the T5 model designed for handling a wide range of natural language processing tasks by framing them as text-to-text problems. It uses an encoder-decoder architecture, which allows it to understand input text and generate corresponding output text, making it highly effective for tasks like translation, summarization, and question answering. One of its key features is its ability to work across 101 languages, as it has been pretrained on a diverse multilingual dataset known as mC4. This makes mT5 particularly relevant for our work, as it can efficiently translate Bengali math word problems into mathematical equations. Additionally, a study~\cite{mt5} demonstrates the potential of mT5 in handling multilingual math word problems, highlighting its capability to generate accurate equations from textual descriptions~\cite{mt5m}. Its strength in both understanding and generating text makes it a strong choice for solving word problems in low-resource languages like Bengali.

\begin{table*}[h]
\centering
\caption{Performance of the proposed models}
\label{tab:my-table111}
\scalebox{1.2}{%
\begin{tabular}{>{\centering\arraybackslash}m{2.0cm}>{\centering\arraybackslash}m{1.8cm}>{\centering\arraybackslash}m{1.8cm}>{\centering\arraybackslash}m{1.8cm}>{\centering\arraybackslash}m{1.8cm}}
\hline
\textbf{Model Name}                   & \textbf{Batch Size}          & \textbf{Epoch} & \textbf{Bleu} & \textbf{Accuracy} \\ \hline
\multirow{6}{*}{Transformer}  & \multirow{3}{*}{8}  & 5     & 68.45  & 47.50\%      \\  
                              &                     & 10    & 82.89  & 72.05\%      \\  
                              &                     & 15    & 87.95  & \textbf{77.30\%}      \\ \cline{2-5} 
                              & \multirow{3}{*}{16} & 5     & 80.75  & 66.00\%      \\  
                              &                     & 10    & 85.28  & 74.05\%      \\  
                              &                     & 15    & 82.18  & 72.35\%      \\ \cline{1-1} \cline{2-5} 
\multirow{6}{*}{mT5}          & \multirow{3}{*}{8}  & 5     & 86.41  & 85.70\%     \\  
                              &                     & 10    & 93.61  & 96.20\%      \\  
                              &                     & 15    & 94.68  & \textbf{97.30\%}      \\ \cline{2-5} 
                              & \multirow{3}{*}{16} & 5     & 40.16  & 14.40\%      \\  
                              &                     & 10    & 93.07  & 95.80\%      \\  
                              &                     & 15    & 94.19  & 96.80\%      \\ \cline{1-1} \cline{2-5} 
\multirow{6}{*}{BanglaT5}          & \multirow{3}{*}{8}  & 5     & 89.14  & 92.40\%    \\ 
                              &                     & 10    & 94.05  & \textbf{95.80\%}      \\  
                              &                     & 15    & 94.06  & 95.50\%      \\ \cline{2-5} 
                              & \multirow{3}{*}{16} & 5     & 93.31  & 95.80\%      \\ 
                              &                     & 10    & 32.50  & 1.30\%     \\  
                              &                     & 15    & 89.96  & 92.20\%      \\ \cline{1-1} \cline{2-5} 
\multirow{1}{*}{mBART50}      & \multirow{1}{*}{8}  & 5     & 95.74  & \textbf{97.20\%}      \\  
                              \hline 
\end{tabular}%
}
\end{table*}

\subsection{BanglaT5}
BanglaT5 is a language-specific variant of the T5 model, specifically tailored for the Bengali language~\cite{bt5}. Like mT5, it uses an encoder-decoder architecture designed to handle text-to-text tasks, making it highly effective for tasks such as translation, summarization, and question answering. BanglaT5 is pretrained on a large Bengali corpus, allowing it to understand the linguistic details and contextual dependencies specific to the Bengali language. This makes it particularly suitable for our work, where the goal is to translate Bengali math word problems into corresponding mathematical equations.

\subsection{mBART-50}
mBART50 is a multilingual sequence-to-sequence transformer model, part of the mBART family, pretrained on data from 50 languages. Like mT5 and BanglaT5, mBART50 utilizes an encoder-decoder architecture, making it highly effective for tasks that involve both understanding and generating text. Its ability to perform translation, summarization, and other generative tasks across multiple languages makes it particularly useful for multilingual environments and low-resource language tasks. For the study, mBART50 is highly relevant because it can effectively handle Bengali, one of the 50 languages in its pretraining. Some studies~\cite{mt5} and~\cite{llm} further highlight the capabilities of multilingual models such as mBART50 in solving math word problems.

\section{\textbf{Experiments}}

\subsection{Baseline setup}

Our work was conducted on Google Colab Notebook with Python 3.10.12, PyTorch 2.0.1, a Tesla T4 GPU (15 GB), 12.5 GB of RAM, and 64 GB of disk space.

\subsection{Evaluation Metrics}

We evaluated our models using two metrics: BLEU score and solution accuracy. The BLEU score measures the similarity between predicted and reference outputs, but it focuses on surface-level matching. This means it may not fully reflect correctness if a model generates the right equation with slight variations in token arrangement, potentially resulting in a lower score. 

\begin{figure}[h]
\centering
\includegraphics[width=1.0\columnwidth]{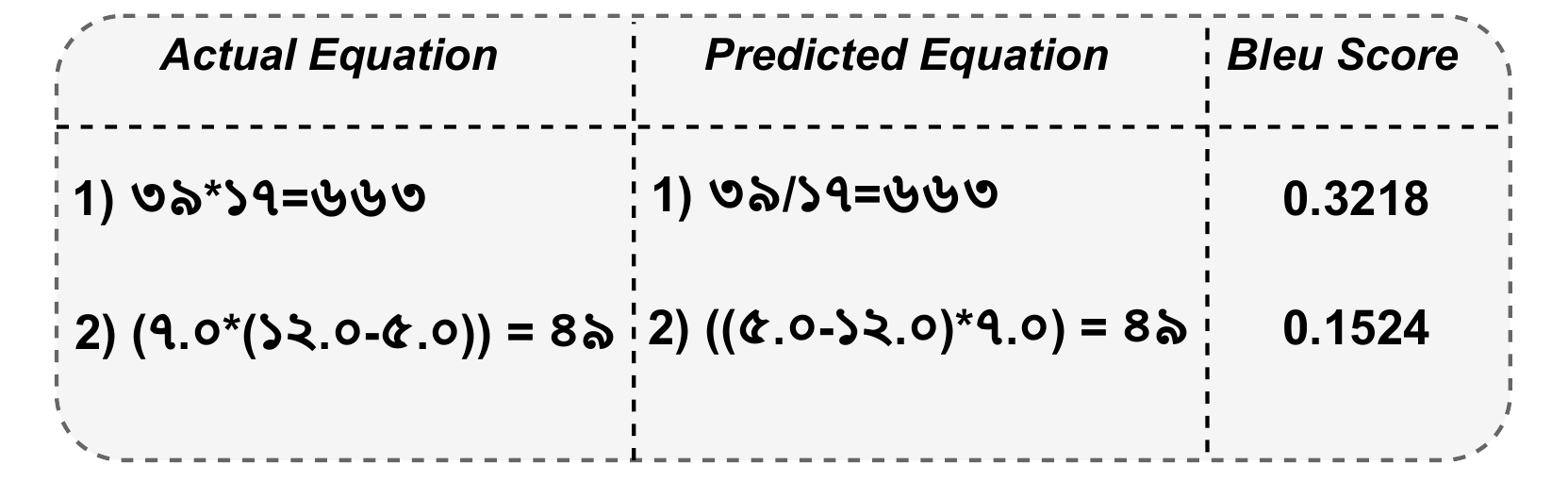} 
\caption{Bleu Score examples}
\label{fig:bleu}
\end{figure}

For instance, in Figure~\ref{fig:bleu} case from our data, the reference equation 2 and the predicted equation 2 had a low BLEU score of 0.1524 despite being mathematically correct. On the other hand the reference equation 1 and the predicted equation 1 had a BLEU score of 0.3218, which is higher than the equation 1, even though it's wrong. This happens because BLEU measures surface-level token similarity rather than evaluating the mathematical equivalence of the equations~\cite{nayak80mathbot}. Hence, minor differences in the order of terms can lead to lower BLEU scores even when the solution is accurate.

To address this, we used solution accuracy, which compares the final solution from the predicted equation with the correct answer. This metric more effectively measures the model's ability to solve math word problems by focusing on the correctness of the numerical result, regardless of minor variations in equation structure.

\subsection{Hyperparameter Tuning}

\begin{table}[h]
\centering
\caption{Hyper Parameters}\label{table1}
\resizebox{\columnwidth}{!}{%
\begin{tabular}{p{3cm}p{3cm}}
\hline
\textbf{Hyperparameter} & \textbf{Enforced Variations} \\
\hline
Dropout & 0.1 \\
Batch Size & 8, 16 \\
Epoch & 5, 10, 15 \\
Learning Rate & 1e-4 \\
\hline
\end{tabular}%
}
\end{table}

Table~\ref{table1} presents the hyperparameters we applied across all models during training. The selection of hyperparameters was based on standard practices and preliminary experimentation to optimize the performance of the model while considering resource constraints. Dropout was fixed at 0.1 to prevent overfitting, as it is commonly effective in transformer models. For batch size, we tested 8 and 16 to balance memory usage and training efficiency, while the number of epochs was varied (5, 10, 15) to capture the effect of extended training. The learning rate was set at 1e-4, a standard value for stable convergence in transformer-based architectures. For mBART50, however, we were constrained to a batch size of 8 with 5 epochs due to resource limitations in Google Colab. Despite this limitation, we maintained consistency in other hyperparameters across all models to ensure a fair comparison.

\section{\textbf{Result Analysis}}

The performance analysis of the models as shown in Table~\ref{tab:my-table111} indicates that mBART50 achieved the highest accuracy of 97.20\% with a BLEU score of 95.74, using a batch size of 8 and 5 epochs. This demonstrates mBART50's strong capability in handling Bengali math word problems. MT5 also performed exceptionally well, reaching a maximum accuracy of 97.30\% and a BLEU score of 94.68 with a batch size of 8 and 15 epochs. BanglaT5 showed competitive performance, with its best accuracy at 95.80\% and a BLEU score of 94.06 using a batch size of 8 and 15 epochs. However, when the batch size was increased to 16, BanglaT5’s accuracy drastically dropped, indicating sensitivity to hyperparameter settings. The Transformer model, while showing gradual improvement with more epochs and larger batch sizes, reached a peak accuracy of 77.30\%, which is significantly lower than that of the other transformer-based models. This result highlights the superior performance of mBART50, MT5, and BanglaT5 over the basic Transformer model in accurately solving Bengali math word problems.

mT5 achieved the highest accuracy among all the models, with a peak performance of 97.30\% accuracy, and a BLEU score of 94.68 when using a batch size of 8 and 15 epochs closely followed by mBART50 which has the accuracy of 97.20\%. This demonstrates mT5's superior capability in translating Bengali math word problems into accurate equations. Its performance remained consistently high across different configurations, indicating its robustness and effectiveness in handling the linguistic complexity of Bengali text. These results show that mT5 is exceptionally well-suited for this task, outperforming other models in terms of overall accuracy.

We proposed the ``PatiGonit" dataset, which has not been the subject of any prior research so far.

\balance
\section{\textbf{Conclusion and Future Works}}

This study presents a novel approach to solving Bengali math word problems using transformer-based models, specifically mT5, BanglaT5, and mBART50. Through the creation of the ``PatiGonit" dataset and the fine-tuning of these models, we successfully demonstrated their ability to translate complex Bengali word problems into accurate mathematical equations. The results show that mT5 outperformed the other models with the highest accuracy of 97.30\%, closely followed by mBART50 and BanglaT5. These findings highlight the effectiveness of transformer models in handling low-resource languages like Bengali, offering significant improvements over traditional deep learning methods. Our work not only advances the field of Bengali natural language processing but also contributes valuable resources and methodologies for future research in educational AI tools. While the research was successful, it faced challenges in translating and culturally adapting math problems into Bengali, which required careful linguistic adjustments. The study also identified a limitation in handling more complex, multi-step problems, suggesting the need for further dataset expansion to enhance the model's robustness and applicability.

\section{\textbf{Limitations}}

Our study faced notable limitations that impacted the scope and effectiveness of the research. Translating English math word problems into Bengali proved challenging due to differences in cultural context, including names and references, which often led to ambiguity or inaccuracies in interpretation. Moreover, the dataset predominantly featured basic arithmetic problems, which constrained the models’ ability to tackle more advanced queries involving complex equations with multiple operators. This lack of diversity in problem types limited the models’ capacity to demonstrate their full potential in addressing a broader range of mathematical challenges.


\begin{thebibliography}{00}
\bibitem{patel2021nlp}
Patel, Arkil, Satwik Bhattamishra, and Navin Goyal. "Are NLP models really able to solve simple math word problems?." arXiv preprint arXiv:2103.07191 (2021).
\bibitem{vaswani2017attention}
Vaswani, A. "Attention is all you need." Advances in Neural Information Processing Systems (2017).
\bibitem{bhattacharjee2021banglabert}
Bhattacharjee, Abhik, et al. "BanglaBERT: Language model pretraining and benchmarks for low-resource language understanding evaluation in Bangla." arXiv preprint arXiv:2101.00204 (2021).
\bibitem{nayak80mathbot}
Nayak, Anish Kumar, Rajeev Patwari, and Viswanathan Subramanian. "MathBot–A Deep Learning based Elementary School Math Word Problem Solver." https://www. semanticscholar. org/paper/MathBot-
\bibitem{feigenbaum2003some}
Feigenbaum, Edward A. "Some challenges and grand challenges for computational intelligence." Journal of the ACM (JACM) 50.1 (2003): 32-40.
\bibitem{7}
Zhang, Biao, et al. "Variational neural machine translation." arXiv preprint arXiv:1605.07869 (2016).
\bibitem{8}
Wang, Yan, Xiaojiang Liu, and Shuming Shi. "Deep neural solver for math word problems." Proceedings of the 2017 conference on empirical methods in natural language processing. 2017.
\bibitem{12}
Chiang, Ting-Rui, and Yun-Nung Chen. "Semantically-aligned equation generation for solving and reasoning math word problems." arXiv preprint arXiv:1811.00720 (2018).
\bibitem{15}
Liang, Zhenwen, et al. "Mwp-bert: Numeracy-augmented pre-training for math word problem solving." arXiv preprint arXiv:2107.13435 (2021).
\bibitem{16}
Raiyan, Syed Rifat, et al. "Math word problem solving by generating linguistic variants of problem statements." arXiv preprint arXiv:2306.13899 (2023).
\bibitem{9}
Wang, Lei, et al. "Translating a math word problem to an expression tree." arXiv preprint arXiv:1811.05632 (2018).
\bibitem{10}
Xie, Zhipeng, and Shichao Sun. "A goal-driven tree-structured neural model for math word problems." Ijcai. 2019.
\bibitem{17}
Chatterjee, Oishik, et al. "WARM: A Weakly (+ Semi) Supervised Model for Solving Math word Problems." arXiv preprint arXiv:2104.06722 (2021).
\bibitem{14}
Qin, Jinghui, et al. "Semantically-aligned universal tree-structured solver for math word problems." arXiv preprint arXiv:2010.06823 (2020).
\bibitem{mt5}
Niyarepola, Kashyapa, et al. "Math word problem generation with multilingual language models." Proceedings of the 15th International Conference on Natural Language Generation. 2022.
\bibitem{mt5m}
Xue, L. "mt5: A massively multilingual pre-trained text-to-text transformer." arXiv preprint arXiv:2010.11934 (2020).
\bibitem{bt5}
Bhattacharjee, Abhik, et al. "BanglaNLG and BanglaT5: Benchmarks and resources for evaluating low-resource natural language generation in Bangla." arXiv preprint arXiv:2205.11081 (2022).
\bibitem{mbart50}
Tang, Yuqing, et al. "Multilingual translation with extensible multilingual pretraining and finetuning." arXiv preprint arXiv:2008.00401 (2020).
\bibitem{llm}
Luo, Liangchen, et al. "Improve Mathematical Reasoning in Language Models by Automated Process Supervision." arXiv preprint arXiv:2406.06592 (2024).
\end{thebibliography}
\end{document}